\documentclass{article}
\usepackage{spconf,amsmath,graphicx}
\usepackage{multirow}
\usepackage{booktabs}
\usepackage{amsfonts}
\usepackage{subfig}
\usepackage{stfloats}
\usepackage{CJKutf8}
\usepackage{xcolor}
\usepackage{amsmath}
\usepackage{adjustbox}
\usepackage{graphicx}
\usepackage{stackengine} 
\usepackage{mathtools}
\usepackage{hyperref}
\usepackage{url}

\title{GSA-TTS: Toward Zero-Shot Speech Synthesis based on Gradual Style Adaptor}

\name{Seokgi Lee\textsuperscript{*}, Jungjun Kim\textsuperscript{*}
\thanks{\textsuperscript{*}Equal contribution.}}
\address{}

\begin{document}
\begin{CJK*}{UTF8}{gbsn}
\maketitle

\begin{abstract}
We present the gradual style adaptor TTS (GSA-TTS) with a novel style encoder that gradually encodes speaking styles from an acoustic reference for zero-shot speech synthesis. GSA first captures the local style of each semantic sound unit. Then the local styles are combined by self-attention to obtain a global style condition. This semantic and hierarchical encoding strategy provides a robust and rich style representation for an acoustic model. We test GSA-TTS on unseen speakers and obtain promising results regarding naturalness, speaker similarity, and intelligibility. Additionally, we explore the potential of GSA in terms of interpretability and controllability, which stems from its hierarchical structure. Audio samples are available at \href{https://gsattsdemo.github.io/}{https://gsattsdemo.github.io/}

\end{abstract}
\section{Introduction}

Over the past few years, the domain of Text-to-Speech (TTS) has witnessed dramatic breakthroughs, driven by the advances in deep learning ~\cite{wang2017tacotron, ren2019fastspeech, kim2020conan, kim2021visual}. In parallel, advancements in grapheme-to-phoneme (G2P) conversion has contributed to more accurate and linguistically informed phonetic representations, further improving the quality of for speech synthesis~\cite{park2020g2pm, kim2023good, yao2015sequence}.

The nascent neural TTS models synthesize sophisticated speech through autoregressive generation. However, autoregressive TTS models suffer from slow inference and weak robustness such as skipping words and repeating words under various conditions. Non-autoregressive TTS models are designed to model duration and time-variant audio variance (e.g., pitch and energy), simplifying one-to-many mapping and facilitating parallel training and inference. Extending to multi-speaker speech synthesis, most studies use speaker-dependent fixed embedding vectors that are jointly trained via backpropagation. As a result, the embedding vector represents the speaker variations. In order to obtain more diverse speaker styles along with speaking contents, the flexible use of reference audio is explored to transfer style attributes. For style adaptation, adaptive instance normalization (AdaIN)~\cite{adain} is used as a popular method that transfers the style to the target domain space using affine transformation. Furthermore, recent studies on the domain of speech synthesis, such as dynamic style adaptation~\cite{pvae} and the hierarchical style generator~\cite{hiervst}, have been proposed for universal speech style transfer.

There is a growing trend toward zero-shot TTS (zs-TTS) with the demand for a custom voice TTS system. A central challenge of zs-TTS is to acquire generalizable style information of an unseen speaker, including global speaker identity and diverse prosody depending on the uttered content. Additionally, the zs-TTS system encounters a discrepancy between reference audio used during training and inference, wherein the undesired content leads to a flaw during inference, often manifesting as blurring or missing words in synthetic audio. In response to the two-fold challenges, a clustering methods like learnable dictionary encoding (LDE)~\cite{LDE}, global style tokens (GST)~\cite{GST}, and x-vector (TDNN)~\cite{TDNN} are employed for general speaker representation while information bottleneck~\cite{inforbottle}, mix-style layer normalization (MSLN)~\cite{MSLN} and dynamic mix-style layer normalization (M-DSLN)~\cite{MDSLN} are proposed to mitigate the superfluous use of the content for style.

Toward the high-fidelity and generalizable zero-shot style transfer, we propose a gradual style adaptor-based text-to-speech model (GSA-TTS), including local style encoder and global style encoder. We introduce a style segmentation strategy for the first time, employing a pre-trained automatic speech recognition (ASR) model. ASR-based segments demonstrate the following fundamental aspects. (1) cutting out non-speech components with respect to style transfer. This ensures that the segments are noise-free and styles are rich with respect to speaker-related information. (2) the output segments represent acoustic units. Interestingly, in our preliminary experiments, we observed that the well-structured segments were exhibited even when the ASR failed, enabling robust style extraction. To be detailed in the segmentation process, we segment the reference mel-spectrogram based on predicted text intervals from ASR. Subsequently, each segmented mel-spectrogram is fed into the local style encoder to represent the compact and prosody-rich style. We expect that each output of the local style encoder contains diverse styles such as stress, intonation, voice identity and linguistic content, represented as a single embedding. Finally, the global style encoder calculates the contributions of each local style followed by averaging the weighted values over time to obtain a fine-grained global style embedding. Based on our experiments, we experienced that gradual style encoding lessens content leakage that arises when using reference audio with excessively mismatched content. Also, gradual style encoding imparts multi-level style features to the acoustic model, resulting in dynamic and high-quality speech. Furthermore, we explore the interpretability and controllability of our gradual style adaptor by analyzing and adjusting the attention weights of the global-style encoder.

\section{Our Method}
We present our proposed model GSA-TTS, which extends upon FastPitch~\cite{fastpitch} that includes online aligner~\cite{online_aligner}, pitch predictor and duration predictor. We introduce a gradual style encoding method to obtain multi-level style conditions. GSA consists of two components: a local-style encoder (LSE) and a global-style encoder (GSE). In the following section, we describe the local style extraction strategy based on ASR model.

\begin{figure*}[t]
\includegraphics[width=\linewidth,clip]{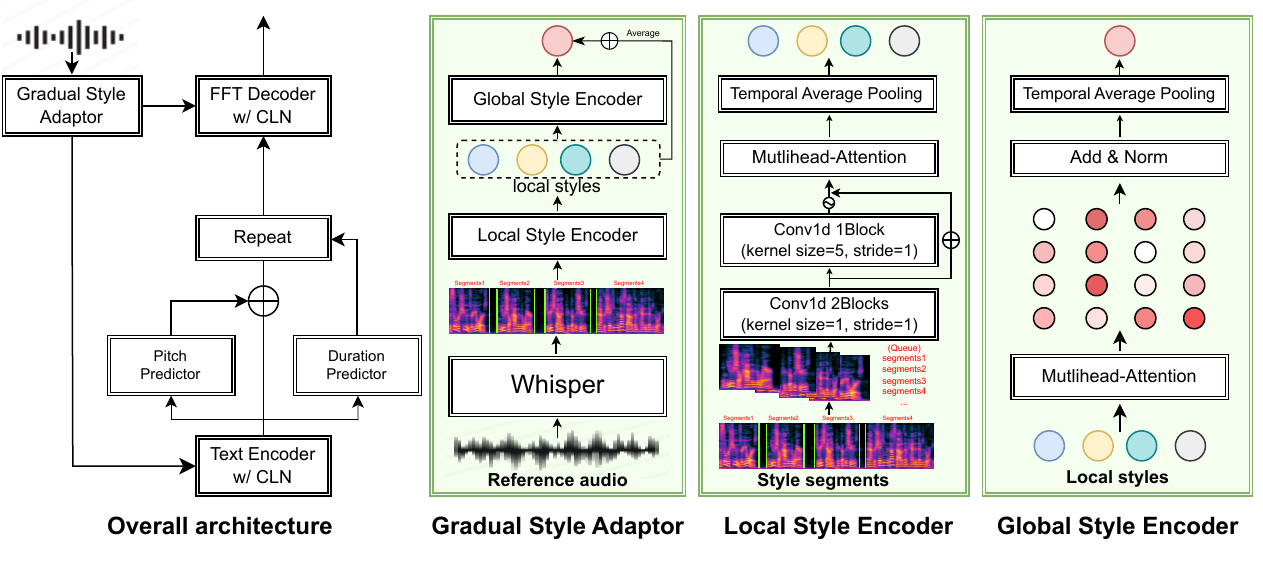}
    \caption{Overall architecture of GSA-TTS.}    
    \label{fig:overall}
\end{figure*}

\subsection{Style Segmentation Strategy}
We set Whisper~\cite{whisper} as our ASR model, which has been trained on 680K hours across 96 languages and achieved state-of-the-art results in many languages and tasks. To determine word-level timesteps, we use the dynamic-time-warping (DTW) method on cross-attention weights derived from aligning audio and text. 

Subsequently, we segment a reference audio by the timesteps. We transform each sliced audio into mel-spectrograms, denoted as style segments. As discussed in ~\cite{intelli}, non-speech frames are redundant to speaker-related styles, degrading speaker identity in synthesized audio. Accordingly, we anticipate that the style segments, being both style-compact and devoid of noise, facilitates the LSE to explicitly capture word-level speaker styles. 

\subsection{Local Style Encoder}
The local-style encoder (LSE) receives the queued style segments as input, aiming to extract a vector containing the word-level style information. Following ~\cite{metastyle}, LSE consists of spectral processing, temporal processing and multihead-attention with temporal average pooling. The spectral processor takes a style segment and transforms it into high-dimensional hidden sequence vectors. In temporal processing, the gated-CNN~\cite{gatedcnn} with a residual connection is used to capture temporal correlations at the frame-level. Then, multihead-attention and temporal average pooling are applied to aggregate spatio-temporal information. We denote an output vector as a local style.

\subsection{Global Style Encoder}
As outlined in \cite{approxitransformer}, self-attention based on Transformer~\cite{transformer} demonstrates a superior ability to act as a universal approximator for permutation-equivalent function. This enables self-attention to robustly handle a variety of inputs for contextual mapping. Commonly, a pairwise dot-product attention is used to weigh the contribution of all latents equally. A weighted combination is utilized to derive contextual embeddings. 

Motivated by the representation power of Transformer, we design our global-style encoder (GSE) based on Transformer with self-attention. The GSE architecture comprises a self-attention layer, a token-wise feed-forward layer, each incorporating two layers with residual connections, followed by a temporal average pooling layer. Through self-attention, the contributions of local styles towards the global style are computed. The contextualized embeddings are averaged over the temporal dimension, giving in a multi-level global vector. Furthermore, to reinforce the time-invariant speaker identity, we average all local styles and add them into the GSE output as a complementary feature.

\subsection{Baseline}
We adopt the FastPitch as our baseline of GSA-TTS. FastPitch integrates encoder and decoder based on feed-forward network, online aligner, pitch predictor and duration predictor. The online aligner not only learns the alignments between text and mel-spectrogram in a monotonic fashion, but also facilitates the efficient training without explicit duration information. The duration predictor is jointly trained with TTS model. 

Following \cite{adaspeech}, conditional layer normalization (CLN) is adopted instead of layer normalization~\cite{layernorm} in each self-attention layer and feed-forward network to condition the global style embedding from GSA. We apply CLN in both the encoder and decoder. CLN linearly projects the normalized input text features into adaptive speaker embedding space with the global style condition. Here, two linear layers denoted as ${E^{\gamma}}$, ${E^{\beta}}$ receive the global style embedding $w$ as input and produce the scale and bias vector, respectively as output.
\begin{equation}
\begin{split}
&CLN(x, w) = \gamma(w) \cdot \frac{x-\mu}{\sigma} + \beta(w),\\
& \gamma(w) = E^{\gamma} * w, \ \ \ \ \ \beta(w) = E^{\beta} * w,
\end{split}
\end{equation}
\noindent where $x$ denotes the input text features. $\mu$ and $\sigma$ are mean and standard deviation along the feature dimension.

\begin{table*}[!t]
\renewcommand{\arraystretch}{1.2}
\setlength{\tabcolsep}{1.6\tabcolsep}
\centering{}

\begin{tabular}{l|cc|ccc}
\hline
\textbf{Model} & \textbf{MOS(↑)}            & \textbf{SMOS(↑)}       & \textbf{SECS(↑)}    & \textbf{WER(↓)}      & \textbf{CER(↓)}          \\ \hline \hline
GT                     & 3.85 ± 0.12    & 3.87 ± 0.11   & 0.795 ± 0.59 & 3.79 ± 0.75     & 1.73 ± 0.36        \\
GT(voc.)                & 3.84 ± 0.11    & 3.72 ± 0.13  & 0.779 ± 0.57 & 4.32 ± 0.86     & 1.98 ± 0.41        \\ \hline\hline
MetaStyleSpeech        & 3.62 ± 0.12    & 3.70 ± 0.12   & 0.613 ± 0.53  & 3.98 ± 0.81     & 2.04 ± 0.46       \\
YourTTS                & 3.66 ± 0.11    & 3.64 ± 0.14   & 0.730 ± 0.51  & 8.90 ± 1.25     & 4.41 ± 0.65       \\ \hline \hline
GSA-TTS(ours)          & \textbf{3.67 ± 0.12}    & \textbf{3.81 ± 0.12} & \textbf{0.792 ± 0.42} & \textbf{1.47 ± 0.46} & \textbf{0.62 ± 0.21}       \\ \hline
\end{tabular}
\caption{Evaluation results regarding the naturalness, speaker identity and intelligibility. \label{table:compare_model}}
\end{table*}

\begin{table*}[!h]
\renewcommand{\arraystretch}{1.2}
\setlength{\tabcolsep}{1.6\tabcolsep}
\centering{}

\begin{tabular}{l|cc|ccc|c}
\hline
\textbf{Method}       & \textbf{MOS(↑)}            & \textbf{CSMOS(↑)}  & \textbf{SECS(↑)}         & \textbf{WER(↓)}      & \textbf{CER(↓)}   & \textbf{\# Params.}      \\ \hline \hline
GSA-TTS                     & \textbf{3.67 ± 0.12}    & 0.00    & 0.792 ± 0.42     &\textbf{1.47 ± 0.46}       &\textbf{0.62 ± 0.21} & 58.80M    \\ \hline \hline
    \emph{w/o} LSE  &-0.07 ± 0.11  &-0.51 ± 0.16 & 0.784 ± 0.43 &8.19 ± 0.22 &2.10 ± 0.54  & 54.07M \\
    \emph{w/o} GSE  &-0.08 ± 0.12 &-0.32 ± 0.18 &0.793 ± 0.42 &3.31 ± 0.80 &1.70 ± 0.46 &57.22M \\
    \emph{w/o} Style Seg. &-0.12 ± 0.12 &-1.26 ± 0.23 &0.701 ± 0.57 &9.03 ± 0.36 &3.44 ± 0.15  &58.80M \\ \hline \hline
FastPitch+MSE[2]  &-0.14 ± 0.12  &-0.78 ± 0.18 & \textbf{0.803 ± 0.44} &15.04 ± 1.76  &8.09 ± 1.06 & 56.73M \\ \hline
\end{tabular}

\caption{MOS, CSMOS, SECS, WER, and CER of an ablation study. Style Seg refers to the style segments. \label{tabel:ablation}}
\end{table*}

\section{Experimental Setup}
We train GSA-TTS using two multi-speaker datasets: LibriTTS-R~\cite{librittsr} (2456 speakers) and the VCTK dataset~\cite{vctk} (109 speakers). For testing, we exclude 11 speakers (p261, p225, p294, p347, p238, p234, p248, p335, p245, p326, and p302) and collect 654 unused transcripts from the VCTK dataset for unseen evaluation. We resample the audio to 22050 kHz sampling rate to ensure compatibility with model training for both datasets. Additionally, in subjective evaluation, we normalize the loudness of all audios to -27 dB for fair comparison. 

For Speaker Encoder Cosine Similarity (SECS), we measure the cosine similarity between the speaker embeddings in synthetic audio and reference audio, resulting in a score ranging from -1 to 1. We employ the speaker encoder from Resemblyzer~\cite{resemblyzer}. For Character Error Rate (CER) and Word Error Rate (WER) evaluation, we use NeMo\footnote{https://github.com/NVIDIA/NeMo} framework. We crowdsourced our subjective evaluation tests through Amazon Mechanical Turk\footnote{https://www.mturk.com/} with the 20 native speakers. We follow the scoring policy for subjective evaluation described in GenerSpeech~\cite{generspeech}.

The GSA-TTS is trained on 4 NVIDIA A100 80GB GPUs for around 150 epochs. We use Adam optimizer with hyperparameters of $\beta_{1}=0.9$, $\beta_{2}=0.98$. Additionally, we employ the learning-rate scheduler designed by \cite{transformer} with a warm-up step count of 4000. The model configurations of FastPitch are unchanged unless otherwise specified. We set the dimension of global style embedding as 384. We employ the HiFi-GAN~\cite{hifigan} vocoder to synthesize the waveform from mel-spectrogram.

\section{Results}
\subsection{Performance}
We compare the performance of naturalness, voice quality and similarity with comparative models using metrics including MOS (mean opinion score), SMOS (similarity mean opinion score), SECS, WER and CER. To ensure a fair comparison, we select open source models as our benchmarks. Our comparative experiment was designed to be straightforward as our GSA seamlessly integrates with existing TTS models in a plug-and-play manner. To evaluate the robustness of GSA-TTS, we synthesize the audio via non-parallel style transfer that uses a random audio from a target speaker as acoustic reference. As shown in Table \ref{table:compare_model}, we confirm statistically significant differences between GSA-TTS and comparative models in CSMOS with the Wilcoxon signed rank test, and in SECS, WER, CER with Wilcoxon rank sum test, at p-level p\texttt{<<}0.05, while there were no statistically significant differences in MOS and SMOS. As MOS and SMOS prioritize assessing naturalness and similarity in a given audio rather than relative comparison, the evaluation of superiority may have been somewhat overlooked.

\begin{figure*}[t]
    \centering
\includegraphics[width=15cm,height=7.0cm,,clip]{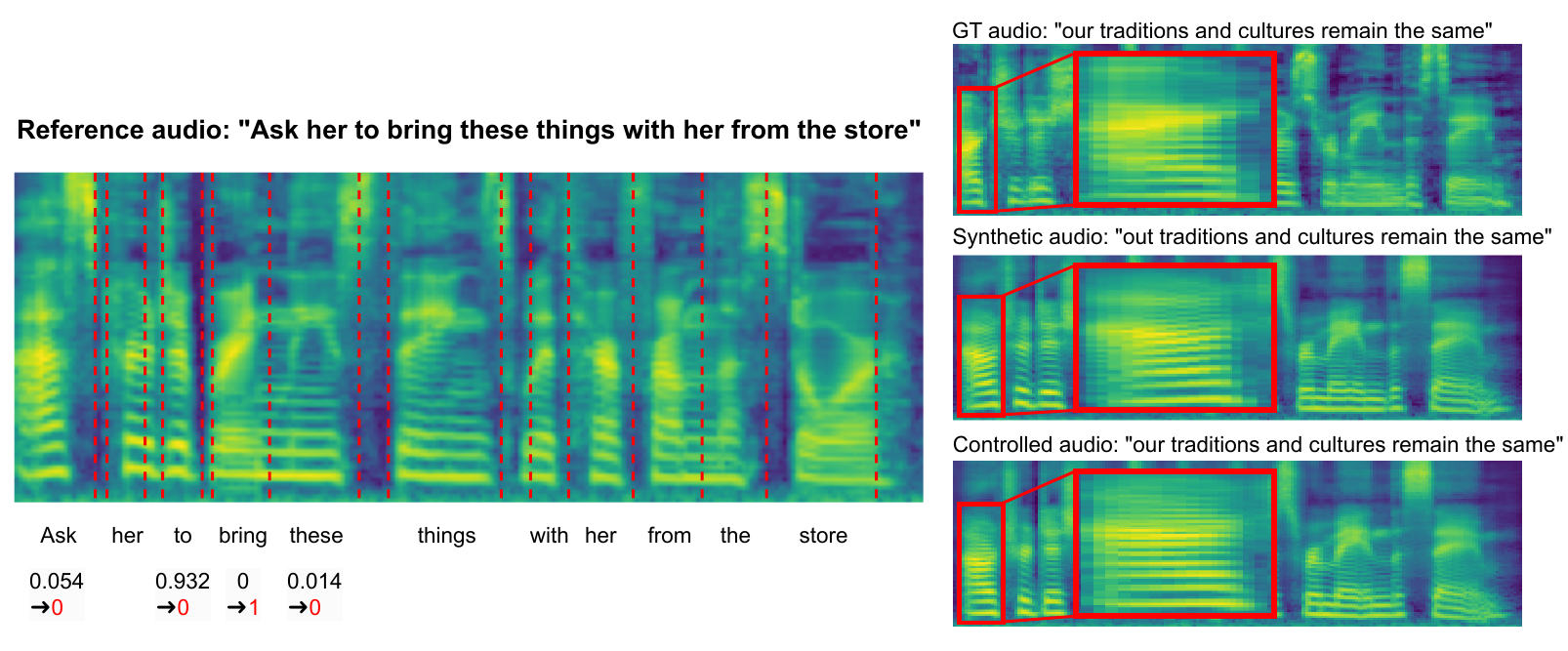}
    \caption{Case study on controllability of GSA-TTS.}    
    \label{fig:casestudy}
\end{figure*}

\begin{figure}[t]
    \centering
\includegraphics[width=5.0cm,height=4.0cm, clip]{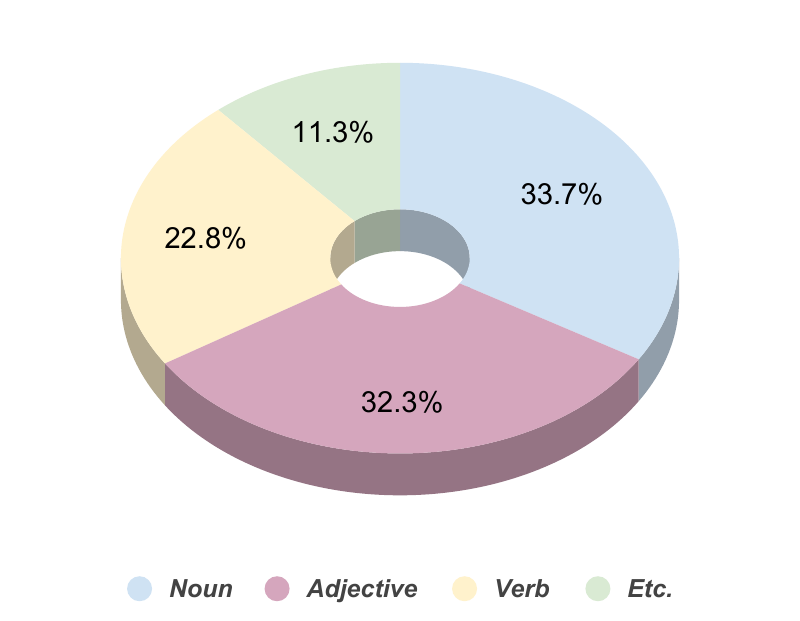}
    \caption{Statistics of highest attention weight on POS tagging.}    
    \label{fig:chart}
\end{figure}

\subsection{Ablation Study}
To validate the effects of LSE, GSE, style segment and GSA, we performed an ablation study using MOS, CSMOS (comparative similarity mean opinion score), SECS, WER and CER tests. Especially, to examine the effect of GSA, we substitute it with a Mel-style encoder (MSE), where CLN is used in both the encoder and decoder to condition the style for fair comparison. We randomly slice the mel spectrogram, maintaining at least 40 frames, and regard each sliced mel-spectrogram as a style segment for the \emph{w/o} Style Seg. experiment.

Table \ref{tabel:ablation} shows the results of an ablation study. As demonstrated, each design contributes to enhancing the quality of GSA-TTS and shows the following findings. The style segment largely enhances the quality of speaker identity, as evidenced by an increase in the CSMOS of 1.26 and the SECS of 0.091. In addition, LSE significantly contributes to the reduction of the content leakage problem, as indicated by the decrease in WER of 6.7 and CER of 1.48. 

Compared to MSE, GSA outperforms overall evaluation metrics by a large margin excluding SECS. The results demonstrate that GSA is more comparative style encoder than MSE in terms of the naturalness, speaker identity and intelligibility. We analyze that the blurred utterances of synthesized audio from MSE attribute the increase of SECS in that the CSMOS of GSA-TTS is remarkably lager than the MSE one.

\subsection{Interpretability and Controllability}
This section explored the interpretability and controllability of GSA-TTS. To find the impact of style segment on speaker similarity and intelligibility, we examined the statistics of the highest attention weight according to the POS (Part of Speech) tagging. As depicted in Figure \ref{fig:chart}, nouns constitute the predominant POS tagging, accounting for 33.7\%. Adjectives closely follow at 32.3\%, while verbs account for 22.8\% and the remainder (Etc.) accounts for 11.3\%. Based on the above observations, we evaluate WER, CER and SECS by modifying the attention weight, where we set the attention weight to zero for all components except the target POS. As presented in Table \ref{table:postag_result}, the style segment tagged with adjectives exhibits the most significant impact on intelligibility while maintaining speaker similarity. Specifically, there is a significant decrease in WER of 0.45. Furthermore, we conduct an empirical analysis of the ratio of voiced frames based on POS tagging. Table \ref{table:voicedframegcratio} illustrates the voiced frame ratio of the style segment based on verbs, adjectives, nouns, and others, resulting in percentages of 73.45\%, 65.73\%, 59.72\%, and 54.68\%, respectively. Combining the above results, we speculate that the style segment tagged with noun, adjective, and verb is more informative compared to other parts of speech tagging.

We further conduct a case study to explore the controllability of GSA-TTS, plotting the mel-spectrograms of reference audio, GT audio, synthetic audio, and controlled audio in Figure \ref{fig:casestudy}. By adjusting the attention weight of the verb 'bring' to 1 and setting the others to zero, we observe a distinct prosodic style tendency between the synthetic audio and the controlled audio. Consequently, controlled audio accurately conveys ``our traditions and cultures remain the same'' mirroring the GT audio, while synthetic audio produces the speech ``out traditions and cultures remain the same''. The ASR result also supports the accuracy of the controlled audio.

\begin{table}[!t]
\renewcommand{\arraystretch}{1.2}
\setlength{\tabcolsep}{1.6\tabcolsep}
\centering{}

\begin{tabular}{l|ll}
\hline
\textbf{POS tagging}      & \textbf{WER(↓)}      & \textbf{SECS(↑)}      \\ \hline \hline
GSA-TTS                  &1.47 ± 0.46       & \textbf{0.792 ± 0.42}             \\ \hline \hline
    Noun    &+0.11 ± 0.52  & 0.791 ± 0.43  \\    
    Verb    &-0.1 ± 0.46 &0.789 ± 0.43  \\
    Adjective    &\textbf{-0.45 ± 0.40} &0.790 ± 0.41  \\
    Noun+Verb+Adjective   &+0.08 ± 0.44 &0.792 ± 0.42 \\ \hline 
\end{tabular}

\caption{Results on WER and SECS according to the attention adjustment with respect to the POS tagging. \label{table:postag_result}}
\end{table}

\begin{table}[t]
\centering
\resizebox{\columnwidth}{!}{
\begin{tabular}{c|c|c|c|c}  \hline                                                 & Verb   & Adjective & Noun   & Etc.   \\ \hline \hline
\begin{tabular}[c]{@{}l@{}}\multicolumn{1}{c}{Voiced}\\ Frame Ratio\end{tabular} & 0.7345 & 0.6573    & 0.5972 & 0.5468 \\ \hline
\end{tabular}
}
\caption{Voiced frame ratio categorized by POS. \label{table:voicedframegcratio}}
\end{table}
\vspace{-0.1cm}

\section{Conclusion}
This paper introduces a novel gradual style adaptor in zero-shot speech synthesis. Throughout the experiments, GSA-TTS shows superior audio quality in terms of naturalness, speaker similarity, and intelligibility compared to existing zero-shot models. In particular, the use of ASR to split local-style units greatly contributes to the performance of GSA-TTS. In addition, GSA can be easily applied to any existing neural TTS model in a plug-and-play manner.

\bibliographystyle{IEEEbib}

\begin{thebibliography}{10}

\bibitem{wang2017tacotron}
Yuxuan Wang, RJ~Skerry-Ryan, Daisy Stanton, Yonghui Wu, Ron~J Weiss, Navdeep Jaitly, Zongheng Yang, Ying Xiao, Zhifeng Chen, Samy Bengio, et~al.,
\newblock ``Tacotron: Towards end-to-end speech synthesis,''
\newblock {\em arXiv preprint arXiv:1703.10135}, 2017.

\bibitem{ren2019fastspeech}
Yi~Ren, Yangjun Ruan, Xu~Tan, Tao Qin, Sheng Zhao, Zhou Zhao, and Tie-Yan Liu,
\newblock ``Fastspeech: Fast, robust and controllable text to speech,''
\newblock {\em Advances in neural information processing systems}, vol. 32, 2019.

\bibitem{kim2020conan}
Jungjun Kim, Hanbin Ko, and Jialin Wu,
\newblock ``Conan: A complementary neighboring-based attention network for referring expression generation,''
\newblock in {\em Proceedings of the 28th International Conference on Computational Linguistics}, 2020, pp. 1952--1962.

\bibitem{kim2021visual}
Jung-Jun Kim, Dong-Gyu Lee, Jialin Wu, Hong-Gyu Jung, and Seong-Whan Lee,
\newblock ``Visual question answering based on local-scene-aware referring expression generation,''
\newblock {\em Neural Networks}, vol. 139, pp. 158--167, 2021.

\bibitem{park2020g2pm}
Kyubyong Park and Seanie Lee,
\newblock ``g2pm: A neural grapheme-to-phoneme conversion package for mandarin chinese based on a new open benchmark dataset,''
\newblock {\em arXiv preprint arXiv:2004.03136}, 2020.

\bibitem{kim2023good}
Jungjun Kim, Changjin Han, Gyuhyeon Nam, and Gyeongsu Chae,
\newblock ``Good neighbors are all you need for chinese grapheme-to-phoneme conversion,''
\newblock in {\em ICASSP 2023-2023 IEEE International Conference on Acoustics, Speech and Signal Processing (ICASSP)}. IEEE, 2023, pp. 1--5.

\bibitem{yao2015sequence}
Kaisheng Yao and Geoffrey Zweig,
\newblock ``Sequence-to-sequence neural net models for grapheme-to-phoneme conversion,''
\newblock {\em arXiv preprint arXiv:1506.00196}, 2015.

\bibitem{adain}
Tero Karras, Samuli Laine, and Timo Aila,
\newblock ``A style-based generator architecture for generative adversarial networks,''
\newblock in {\em Proceedings of the IEEE/CVF conference on computer vision and pattern recognition}, 2019, pp. 4401--4410.

\bibitem{pvae}
Ji-Hyun Lee, Sang-Hoon Lee, Ji-Hoon Kim, and Seong-Whan Lee,
\newblock ``Pvae-tts: adaptive text-to-speech via progressive style adaptation,''
\newblock in {\em ICASSP 2022-2022 IEEE International Conference on Acoustics, Speech and Signal Processing (ICASSP)}. IEEE, 2022, pp. 6312--6316.

\bibitem{hiervst}
Sang-Hoon Lee, Ha-Yeong Choi, Hyung-Seok Oh, and Seong-Whan Lee,
\newblock ``Hiervst: Hierarchical adaptive zero-shot voice style transfer,''
\newblock {\em arXiv preprint arXiv:2307.16171}, 2023.

\bibitem{LDE}
Erica Cooper, Cheng-I Lai, Yusuke Yasuda, Fuming Fang, Xin Wang, Nanxin Chen, and Junichi Yamagishi,
\newblock ``Zero-shot multi-speaker text-to-speech with state-of-the-art neural speaker embeddings,''
\newblock in {\em ICASSP 2020-2020 IEEE International Conference on Acoustics, Speech and Signal Processing (ICASSP)}. IEEE, 2020, pp. 6184--6188.

\bibitem{GST}
Yuxuan Wang, Daisy Stanton, Yu~Zhang, RJ-Skerry Ryan, Eric Battenberg, Joel Shor, Ying Xiao, Ye~Jia, Fei Ren, and Rif~A Saurous,
\newblock ``Style tokens: Unsupervised style modeling, control and transfer in end-to-end speech synthesis,''
\newblock in {\em International conference on machine learning}. PMLR, 2018, pp. 5180--5189.

\bibitem{TDNN}
Cancan Jin, Ben He, Kai Hui, and Le~Sun,
\newblock ``Tdnn: a two-stage deep neural network for prompt-independent automated essay scoring,''
\newblock in {\em Proceedings of the 56th Annual Meeting of the Association for Computational Linguistics (Volume 1: Long Papers)}, 2018, pp. 1088--1097.

\bibitem{inforbottle}
Xudong Dai, Cheng Gong, Longbiao Wang, and Kaili Zhang,
\newblock ``Information sieve: Content leakage reduction in end-to-end prosody transfer for expressive speech synthesis.,''
\newblock in {\em Interspeech}, 2021, pp. 131--135.

\bibitem{MSLN}
Rongjie Huang, Yi~Ren, Jinglin Liu, Chenye Cui, and Zhou Zhao,
\newblock ``Generspeech: Towards style transfer for generalizable out-of-domain text-to-speech,''
\newblock {\em Advances in Neural Information Processing Systems}, vol. 35, pp. 10970--10983, 2022.

\bibitem{MDSLN}
Ji-Hoon Kim, Hong-Sun Yang, Yoon-Cheol Ju, Il-Hwan Kim, and Byeong-Yeol Kim,
\newblock ``Crossspeech: Speaker-independent acoustic representation for cross-lingual speech synthesis,''
\newblock in {\em ICASSP 2023-2023 IEEE International Conference on Acoustics, Speech and Signal Processing (ICASSP)}. IEEE, 2023, pp. 1--5.

\bibitem{fastpitch}
Adrian {\L}a{\'n}cucki,
\newblock ``Fastpitch: Parallel text-to-speech with pitch prediction,''
\newblock in {\em ICASSP 2021-2021 IEEE International Conference on Acoustics, Speech and Signal Processing (ICASSP)}. IEEE, 2021, pp. 6588--6592.

\bibitem{online_aligner}
Rohan Badlani, Adrian {\L}a{\'n}cucki, Kevin~J Shih, Rafael Valle, Wei Ping, and Bryan Catanzaro,
\newblock ``One tts alignment to rule them all,''
\newblock in {\em ICASSP 2022-2022 IEEE International Conference on Acoustics, Speech and Signal Processing (ICASSP)}. IEEE, 2022, pp. 6092--6096.

\bibitem{whisper}
Alec Radford, Jong~Wook Kim, Tao Xu, Greg Brockman, Christine McLeavey, and Ilya Sutskever,
\newblock ``Robust speech recognition via large-scale weak supervision,''
\newblock in {\em International Conference on Machine Learning}. PMLR, 2023, pp. 28492--28518.

\bibitem{intelli}
Sunghee Jung, Won Jang, Jaesam Yoon, and Bongwan Kim,
\newblock ``Intelli-z: Toward intelligible zero-shot tts,''
\newblock {\em arXiv preprint arXiv:2401.13921}, 2024.

\bibitem{metastyle}
Dongchan Min, Dong~Bok Lee, Eunho Yang, and Sung~Ju Hwang,
\newblock ``Meta-stylespeech: Multi-speaker adaptive text-to-speech generation,''
\newblock in {\em International Conference on Machine Learning}. PMLR, 2021, pp. 7748--7759.

\bibitem{gatedcnn}
Yann~N Dauphin, Angela Fan, Michael Auli, and David Grangier,
\newblock ``Language modeling with gated convolutional networks,''
\newblock in {\em International conference on machine learning}. PMLR, 2017, pp. 933--941.

\bibitem{approxitransformer}
Chulhee Yun, Srinadh Bhojanapalli, Ankit~Singh Rawat, Sashank~J Reddi, and Sanjiv Kumar,
\newblock ``Are transformers universal approximators of sequence-to-sequence functions?,''
\newblock {\em arXiv preprint arXiv:1912.10077}, 2019.

\bibitem{transformer}
Ashish Vaswani, Noam Shazeer, Niki Parmar, Jakob Uszkoreit, Llion Jones, Aidan~N Gomez, {\L}ukasz Kaiser, and Illia Polosukhin,
\newblock ``Attention is all you need,''
\newblock {\em Advances in neural information processing systems}, vol. 30, 2017.

\bibitem{adaspeech}
Mingjian Chen, Xu~Tan, Bohan Li, Yanqing Liu, Tao Qin, Sheng Zhao, and Tie-Yan Liu,
\newblock ``Adaspeech: Adaptive text to speech for custom voice,''
\newblock {\em arXiv preprint arXiv:2103.00993}, 2021.

\bibitem{layernorm}
Jimmy~Lei Ba, Jamie~Ryan Kiros, and Geoffrey~E Hinton,
\newblock ``Layer normalization,''
\newblock {\em arXiv preprint arXiv:1607.06450}, 2016.

\bibitem{librittsr}
Yuma Koizumi, Heiga Zen, Shigeki Karita, Yifan Ding, Kohei Yatabe, Nobuyuki Morioka, Michiel Bacchiani, Yu~Zhang, Wei Han, and Ankur Bapna,
\newblock ``Libritts-r: A restored multi-speaker text-to-speech corpus,''
\newblock {\em arXiv preprint arXiv:2305.18802}, 2023.

\bibitem{vctk}
Christophe Veaux, Junichi Yamagishi, Kirsten MacDonald, et~al.,
\newblock ``Superseded-cstr vctk corpus: English multi-speaker corpus for cstr voice cloning toolkit,''
\newblock 2016.

\bibitem{resemblyzer}
Corentin Jemine et~al.,
\newblock ``Real-time voice cloning,''
\newblock 2019.

\bibitem{generspeech}
Rongjie Huang, Yi~Ren, Jinglin Liu, Chenye Cui, and Zhou Zhao,
\newblock ``Generspeech: Towards style transfer for generalizable out-of-domain text-to-speech,''
\newblock {\em Advances in Neural Information Processing Systems}, vol. 35, pp. 10970--10983, 2022.

\bibitem{hifigan}
Jungil Kong, Jaehyeon Kim, and Jaekyoung Bae,
\newblock ``Hifi-gan: Generative adversarial networks for efficient and high fidelity speech synthesis,''
\newblock {\em Advances in neural information processing systems}, vol. 33, pp. 17022--17033, 2020.

\end{thebibliography}

\end{CJK*}
\end{document}